\documentclass[11pt,a4paper]{article}
\usepackage[hyperref]{acl2019}
\usepackage{times}
\usepackage{latexsym}

\usepackage{url}
\usepackage{graphicx}  

\usepackage{pgfplots}
\usepackage{amsmath}
\usepackage{amssymb}
\usepackage{array}

\DeclareMathOperator*{\argmax}{argmax} 

\aclfinalcopy 
\def\aclpaperid{1915} 

\title{Predicting Human Activities from User-Generated Content}

\author{Steven R. Wilson \and Rada Mihalcea\\
  University of Michigan \\
  \texttt{\{steverw,mihalcea\}@umich.edu}}

\date{}

\begin{document}
\maketitle
\begin{abstract}
The activities we do are linked to our interests, personality, political preferences, and decisions we make about the future. In this paper, we explore the task of predicting human activities from user-generated content. We collect a dataset containing instances of social media users writing about a range of everyday activities. We then use a state-of-the-art sentence embedding framework tailored to recognize the semantics of human activities and perform an automatic clustering of these activities.  We train a neural network model to make predictions about which clusters contain activities that were performed by a given user based on the text of their previous posts and self-description. Additionally, we explore the degree to which incorporating inferred user traits into our model helps with this prediction task.
\end{abstract}

\section{Introduction}

What a person does says a lot about who they are. Information about the types of activities that a person engages in can provide insights about their interests \cite{goecks2000learning}, personality \cite{ajzen1987}, physical health \cite{bouchard2018physical}, the activities that they are likely to do in the future \cite{ouellette1998habit}, and other psychological phenomena like personal values \cite{rokeach1973nature}. For example, it has been shown that university students who  exhibit traits of interpersonal affect and self-esteem are more likely to attend parties \cite{paunonen2001big}, and those that value stimulation are likely to watch movies that can be categorized as thrillers \cite{bardi2003values}.

Several studies have applied computational approaches to the understanding and modeling of human behavior at scale \cite{yin2014temporal} and in real time \cite{wang2015smartgpa}. However, this previous work has mainly relied on specific devices or platforms that require structured definitions of behaviors to be measured. While this leads to an accurate understanding of the types of activities being done by the involved users, these methods capture a relatively narrow set of behaviors compared to the huge range of things that people do on a day-to-day basis. On the other hand, publicly available social media data provide us with information about an extremely rich and diverse set of human activities, but the data are rarely structured or categorized, and they mostly exist in the form of natural language. Recently, however, natural language processing research has provided several examples of methodologies for extracting and representing human activities from text \cite{fast2016augur,wilson2017measuring} and even multimodal data \cite{agrawal2016sort}.

In this paper, we explore the task of predicting  human activities from user-generated text data, which will allow us to gain a deeper understanding of the kinds of everyday activities that people discuss online with one another. Throughout the paper, we use the word ``activity'' to refer to what an individual user does or has done in their daily life. Unlike the typical use of this term in the computer vision community \cite{cheng2015advances,zhang2017review}, in this paper we use it in a broad sense, to also encompass non-visual activities such as ``make vacation plans" or ``have a dream'' We do not focus on fine-grained sequences actions such as ``pick up a camera'', ``hold a camera to one's face'', ``press the shutter release button'', and others. Rather, we focus on the high-level activity as a person would report to others: ``take a picture''. Additionally, we specifically focus on everyday human activities done by the users themselves, rather than larger-scale events \cite{atefeh2015survey}, which are typically characterized by the involvement or interest of many users, often at a specific time and location.

Given that the space of possible phrases describing human activities is nearly limitless, we propose a set of human activity clusters that summarize a large set of several hundred-thousand self-reported activities. We then construct predictive models that are able to estimate the likelihood that a user has reported that they have performed an activity from any cluster. 

The paper makes the following main contributions. First, starting with a set of nearly 30,000 human activity patterns, we compile a very large dataset of more than 200,000 users undertaking one of the human activities matching these patterns, along with over 500 million total tweets from these users. Second, we use a state-of-the-art sentence embedding framework tailored to recognize the semantics of human activities and create a set of activity clusters of variable granularity. Third, we explore a neural model that can predict human activities based on natural language data, and in the process also investigate the relationships between everyday human activities and other social variables such as personal values.

\section{Data} 


While we do not expect to know exactly what a person is doing at any given time, it is fairly common for people to publicly share the types of activities that they are doing by making posts, written in natural language, on social media platforms like Twitter. However, when taking a randomly sampled stream of tweets, we find that only a small fraction of the content was directly related to activities that the users were doing in the real world -- instead, most instances are more conversational in nature, or contain the sharing of opinions about the world or links to websites or images. Using such a random sample would require us to filter out a large percentage of the total data collected, making the data collection process inefficient.

Therefore, in order to target only those tweets that are rich in human activity content, we formulate a set of queries that allows us to use the Twitter Search API to find instances of users tweeting about common human activities. Each query contains a first-person, past-tense verb within a phrase that describes a common activity that people do. Using this approach, we are able to retrieve a set of tweets that contains a high concentration of human activity content, and we also find that users who wrote these tweets are much more likely to have written \textit{other} tweets that describe human activities (Table \ref{tab:query_effect}). We build our set of human activity queries from two sources: the Event2Mind dataset \cite{event2mind} and a set of short activity surveys, which we collect ourselves, to obtain nearly 30K queries (Table \ref{tab:query_size}) .

\begin{table}[t]
\centering
\scalebox{0.9}{
\begin{tabular}{ll}
\hline
Sampled tweets w/valid activities                       & 2\% \\ 
Queried tweets w/valid activities                      & 81\% \\ 
Addtl. user tweets w/valid activities & 15\% \\ \hline
\end{tabular}
}
\caption{Effect of targeted query approach on activity frequency in tweets. ``Valid activities'' are defined as first-person verb phrases that clearly indicate that the author of the text has actually performed the concrete activity being described. For each set of tweets, a random subset of 100 was chosen and manually annotated for validity.}
\label{tab:query_effect}
\end{table}

\begin{table}[t]
\centering
\scalebox{0.9}{
\begin{tabular}{lll}
\hline
                     & count      & unique     \\ \hline
Event2Mind activities  & 24,537          & 24,537          \\
Survey activities    & 5,000          & 4,957          \\ \hline
\textbf{Total}       & \textbf{29,537} & \textbf{29,494} \\ \hline
\end{tabular}
}
\caption{Number of human activity queries from multiple sources.}
\label{tab:query_size}
\end{table}

\subsection{Event2Mind Activities} The Event2Mind dataset contains a large number of event phrases which are annotated for intent and reaction. The events themselves come from four sources of phrasal events (stories, common n-grams found in web data, blogs, and English idioms), and many of them fall under our classification of human activities, making Event2Mind a great resource in our search for concrete examples of human activities. We consider events for which a person is the subject (e.g, ``PersonX listens to PersonX's music'') to be human activities, and remove the rest (e.g., ``It is Christmas morning''). We then use several simple rules to convert the Event2Mind instances into first-person past-tense activities. Since all events were already filtered so that they begin with ``PersonX'', we replace the first occurrence of ``PersonX'' in each event with ``I'' and all subsequent occurrences with ``me''. All occurrences of ``PersonX's'' become ``my'', and the main verb in each phrase is conjugated to its past-tense form using the Pattern python module.\footnote{www.clips.uantwerpen.be/pattern} For example, the event ``PersonX teaches PersonX's son'' becomes the query ``I taught my son''. Since Event2Mind also contains wildcard placeholders that can match any span of text within the same phrase (e.g., ``PersonX buys $\_\_\_$ at the store'')\footnote{We also treat instance of ``PersonY'' as a wildcard since this could be any name or even a user (@) mention on Twitter.} but the Twitter API doesn't provide a mechanism for wildcard search, we split the event on the string $\_\_\_$ and generate a query that requires all substrings to appear in the tweet. We then check all candidate tweets after retrieval and remove any for which the substrings do not appear in the same order as the original pattern.

\subsection{Short Survey Activities} In order to get an even richer set of human activities, we also ask a set of 1,000 people across the United States to list any five activities that they had done in the past week. We collect our responses using Amazon Mechanical Turk,\footnote{www.mturk.com} and manually verify that all responses are reasonable. We remove any duplicate strings and automatically convert them into first-person and past-tense (if they were not in that form already). For this set of queries, there are no wildcards and we only search for exact matches. Example queries obtained using this approach include ``I went to the gym'' and ``I watched a documentary''.


\begin{table}[]
\centering
\scalebox{0.9}{
\begin{tabular}{ll}
\hline
Total queries               & 29,494                      \\ 
Queried tweets               & 422,607                      \\ 
Avg. tweets/query          & 14.33                      \\ 
Valid queried tweets               & 335,357                      \\ 
Avg. valid tweets/query               & 11.37                      \\ \hline
\end{tabular}
}
\caption{Summary of query results.}
\label{tab:query_results}
\end{table}


\subsection{Query Results} Using our combined set of unique human activity queries, we use the Twitter Search API\footnote{developer.twitter.com/en/docs/tweets/search/api-reference/get-search-tweets.html} to collect the most recent 100 matches per query (the maximum allowed by the API per request), as available, and we refer to these tweets as our set of \textit{queried tweets}. We then filter the \textit{queried tweets} as follows: first, we verify that for any tweets requiring the match of multiple substrings (due to wildcards in the original activity phrase), the substrings appear in the correct order and do not span multiple sentences. Next, we remove activity phrases that are preceded with indications that the author of the tweet did not actually perform the activity, such as ``I wish'' or ``should I \ldots ?''. We refer to the set of tweets left after this filtering as \textit{valid queried tweets} (see Table \ref{tab:query_results} for more details).

\begin{table}[]
\centering
\scalebox{0.9}{
\begin{tabular}{ll}
\hline
Num. unique users           & 358,091                      \\ \hline
Additional tweets collected                         & 560,526,633 \\ 
Avg. additional tweets / user                         & 1,565 \\ \hline
Additional activities extracted                         & 21,316,364 \\ 
Avg. additional activities / user                       & 59.52 \\ \hline

\end{tabular}
}
\caption{Summary of additional data.}
\label{tab:addtl_size}
\end{table}

\begin{table}[t]
\centering
\scalebox{0.9}{
\begin{tabular}{ll}
\hline
Initial number unique users           & 358,091                      \\ \hline
Users with non-empty profiles      & 96.9\%                      \\ 
Users with $\geq$ 1 addtl. tweets                & 94.9\% \\
Users with $\geq$ 25 addtl. tweets                & 93.1\% \\
Users with $\geq$ 1 addtl. activities    & 93.5\% \\ 
Users with $\geq$ 5 addtl. activities & 87.1\% \\ \hline
\textbf{Final number unique valid users}           & \textbf{214,708}                      \\ \hline
\end{tabular}
}
\caption{Summary valid user filtering.}
\label{tab:user_filtering}
\end{table}

In order to gather other potentially useful information about the users who wrote at least one \textit{valid queried tweet}, we collect both their self-written profile and their previously written tweets (up to 3,200 past tweets per user, as allowed by the Twitter API), and we refer to these as our set of \textit{additional tweets}. We ensure that there is no overlap between the sets of \textit{queried tweets} and \textit{additional tweets}, so in the unlikely case that a user has posted the same tweet multiple times, it cannot be included in both sets. 

Further, we use a simple pattern-matching approach to extract additional activities from these \textit{additional tweets}. We search for strings that match \texttt{I <VBD> .* <EOS>} where \texttt{<VBD>} is any past-tense verb, \texttt{.*} matches any string (non-greedy), and \texttt{<EOS>} matches the end of a sentence. We then perform the same filtering as before for indications that the person did not actually do the activity, and we refer to these filtered matches as our set of \textit{additional activities} (see Table \ref{tab:addtl_size} for more information). Note that since these \textit{additional activities} can contain any range of verbs, they are naturally noisier than our set of \textit{valid query tweets}, and we therefore do not treat them as a reliable ``ground truth'' source of self-reported human activities, but as a potentially useful signal of activity-related information that can be associated with users in our dataset. 

For our final dataset, we also filter our set of users. From the set of users who posted at least one \textit{valid queried tweet}, we remove those who had empty user profiles, those with less than 25 additional tweets, and those with less than 5 additional activities (Table \ref{tab:user_filtering}).

\subsection{Creating Human Activity Clusters}


Given that the set of possible human activity phrases is extremely large and it is unlikely that the same phrase will appear multiple times, we make this space more manageable by first performing a clustering over the set of \textit{activity phrase instances} that we extract from all \textit{valid queried tweets}. We define an \textit{activity phrase instance} as the set of words matching an activity query, plus all following words through the end of the sentence in which the match appears. By doing this clustering, our models will be able to make a prediction about the likelihood that a user has mentioned activities from each cluster, rather than only making predictions about a single point in the semantic space of human activities.


In order to cluster our \textit{activity phrase instances}, we need to define a notion of distance between any pair of instances. For this, we turn to prior work on models to determine semantic similarity between human activity phrases \cite{zhang2018sequential} in which the authors utilized transfer learning in order to fine-tune the Infersent \cite{conneau2017supervised} sentence similarity model to specifically capture relationships between human activity phrases. We use the authors' BiLSTM-max sentence encoder trained to capture the relatedness dimension of human activity phrases\footnote{Shared by the first author of the referenced paper.} to obtain vector representations of each of our activity phrases. The measure of distance between vectors produced by this model was shown to be strongly correlated with human judgments of general activity relatedness (Spearman's $\rho=.722$ between the model and human ratings, while inter-annotator agreement is $.768$). 

\begin{table}[t]
\centering
\scalebox{0.9}{
\begin{tabular}{c} \hline
\textbf{``Cooking''}\\ \hline
make cauliflower stir-fry for dinner\\
make garlic and olive oil vermicelli for lunch\\
start cooking bacon in the oven (on foil in a sheet)\\ 
burn the turkey\\
make perfect swordfish steaks tonight\\ \hline
\textbf{``Pet/Animal related''}\\ \hline
get a new pet spider today \\
cuddle 4 dogs \\
get a pet sitter \\
feel so happy being able to pet kitties today \\
spend some time with cats \\ \hline
\textbf{``Spectating''}\\ \hline
watch football italia \\
watch a football game in the pub \\
watch basketball today \\
watch sports \\
watch fireworks today in the theatre \\ \hline
\textbf{``Passing Examinations''}\\ \hline
ace the exam \\
pass one's exam thank god \\
get a perfect score on one's exam \\
get a c on one's french exam \\
pass another exam omg \\ \hline

\end{tabular}
}
\caption{Examples of clustered activities (with manually provided labels, for reference purposes only).}
\label{tab:cluster_examples}
\end{table}

While the relationship between two activity phrases can be defined in a number of ways \cite{wilson2017measuring}, we we chose a model that was optimized to capture relatedness so that our clusters would contain groups of related activities without enforcing that they are strictly the same activity. Since the model that we employed was trained on activity phrases in the infinitive form, we again use the Pattern python library, this time to convert all of our past-tense activities to this form. We also omit the leading first person pronoun from each phrase, and remove user mentions (\texttt{@<user>}), hashtags, and URLs. We then define the distance between any two vectors using cosine distance, i.e., $1-\frac{\mathbf{A}\cdot \mathbf{B}}{||\mathbf{A}|| ||\mathbf{B}||}$, for vectors $\mathbf{A}$ and $\mathbf{B}$. 


\begin{table}[t]
    \centering
    \scalebox{0.9}{
    \begin{tabular}{c} \hline
         \textbf{Distance to ``Cooking'': 0.11} \\ \hline
         cook breakfast\\
         cook the spaghetti\\
         start cooking\\
         cook something simple\\
         start cooking a lot more \\ \hline
         \textbf{Distance to ``Cooking'': 0.52} \\ \hline
         feed one's ducks bread all the time \\
         give one's dog some chicken \\
         stop eating meat \\
         eat hot dogs and fries \\
         get one's dog addicted to marshmellows \\ \hline
         \textbf{Distance to ``Cooking'': 0.99} \\ \hline
         take a picture with her\\
         post a photo of one\\
         bring something like 1000 rolls of film\\
         draw a picture of us holding hands\\
         capture every magical moment to give to the bride \\ \hline
    \end{tabular}
    }
    \caption{Three sample clusters and their distances from the first cluster in Table \ref{tab:cluster_examples}, showing the closest cluster, a somewhat distant cluster, and a very distant cluster. \vspace{-0.1in}}
    \label{tab:similar_clusters}
\end{table}

We use K-means clustering in order to find a set of $k_{act}$ clusters that can be used to represent the semantic space in which the activity vectors lie. 
We experiment with $k_{act}=2^n$ with $n\in \mathbb{Z} \cap [3,13]$ and evaluate the clustering results using several metrics that do not require supervision: within-cluster variance, silhouette coefficient \cite{rousseeuw1987silhouettes}, Calinski-Harabaz criterion \cite{calinski1974dendrite}, and Davies-Bouldin criterion \cite{davies1979cluster}. In practice, however, we find that these metrics are strongly correlated (either positively or negatively) with the $k_{act}$, making it difficult to quantitatively compare the results of using a different number of clusters, and we therefore make a decision based on a qualitative analysis of the clusters.\footnote{We acknowledge that similar experiments could be run with different cluster assignments, and our preliminary experiments showed comparable results. 
It is important to note that we do not treat these clusters as \textit{the} definitive organization of human activities, but as an approximation of the full activity space in order to reduce the complexity of making predictions about activities in that space.}  For the purpose of making these kinds of predictions about clusters, it is beneficial to have a smaller number of larger clusters, but clusters that are too large are no longer meaningful since they contain sets of activities that are less strongly related to one another. In the end, we find that using $2^{10}=1024$ clusters leads to a good balance between cluster size and specificity, and we use this configuration for our prediction experiments moving forward. Examples of activities that were assigned the same cluster label are shown in Table \ref{tab:cluster_examples}, and Table \ref{tab:similar_clusters} illustrates the notion of distance within our newly defined semantic space of human activities. For example, two cooking-related clusters are near to one another, while a photography-related cluster is very distant from both.

\section{Methodology}

Given a set of activity clusters and knowledge about the users who have reported to have participated in these activities, we explore the ability of machine learning models to make inferences about which activities are likely to be next performed by a user. Here we describe the supervised learning setup, evaluation, and neural architecture used for the prediction task.


\subsection{Problem Statement}

We formulate our prediction problem as follows: for a given user, we would like to produce a probability distribution over all activity clusters such that:
$$
\argmax_{c_i\in C} P(c_i|\mathbf{h},\mathbf{p},\mathbf{a}) = c_t \,,
$$
where $C$ is a set of activity clusters, $\mathbf{h}$, $\mathbf{p}$, and $\mathbf{a}$ are vectors that represent the user's \textbf{history}, \textbf{profile}, and \textbf{attributes}, respectively, and $c_t$ is the target cluster. The target cluster is the cluster label of an activity cluster that contains an activity that is known to have been performed by the user.

If a model is able to accurately predict the target cluster, then it is able to estimate the general type of activity that the user is likely to write about doing in the future given some set of information about the user and what they have written in the past. By also generating a probability distribution over the clusters, we can assign a likelihood that each user will write about performing each group of activities in the future. For example, such a model could predict the likelihood that a person will claim to engage in a ``Cooking'' activity or a  ``Pet/Animal related'' activity.

The ability to predict the exact activity cluster correctly is an extremely difficult task, and in fact, achieving that alone would be a less informative result than producing predictions about the likelihood of all clusters. Further, in our setup, we only have knowledge about a sample of activities that people actually have done. In reality, it is very likely that users have participated in activities that belong to a huge variety of clusters, regardless of which activities were actually reported on social media. Therefore, it should be sufficient for a model to give a relatively high probability to any activity that has been reported by a user, even if there is no report of the user having performed an activity from the cluster with the highest probability for that user.

\subsection{Model Architecture} \label{sec:arch}

As input to our activity prediction model, we use three major components: a user's \textbf{history}, \textbf{profile}, and \textbf{attributes}. We represent a \textbf{history} as a sequence of documents, $D$, written by the user, that contain information about the kinds of activities that they have done. Let $t=|D|$, and each document in $D$ is represented as a sequence of tokens. We experiment with two sources for $D$: all \textit{additional tweets} written by a user, or only the \textit{additional activities} contained in tweets written by a user, which is a direct subset of the text contained in the full set of tweets. 


A user's \textbf{profile} is a single document, also represented as a sequence of tokens. For each user, we populate the \textbf{profile} input using the plain text user description associated with their account, which often contains terms which express self-identity such as ``republican'' or ``athiest.'' 

We represent the tokens in both the user's history and profile with the pretrained 100-dimensional GloVe-Twitter word embeddings \cite{pennington2014glove}, and preprocess all text with the script included with these embeddings.\footnote{nlp.stanford.edu/projects/glove/preprocess-twitter.rb} 

Finally, our model allows the inclusion of any additional \textbf{attributes} that might be known or inferred in order to aid the prediction task, which can be passed to the model as a $dim_a$ dimensional real-valued vector. For instance, we can use personal values as a set of attributes, as described in Section \ref{sec:values}.

\begin{figure}
\centering
\includegraphics[width=7cm]{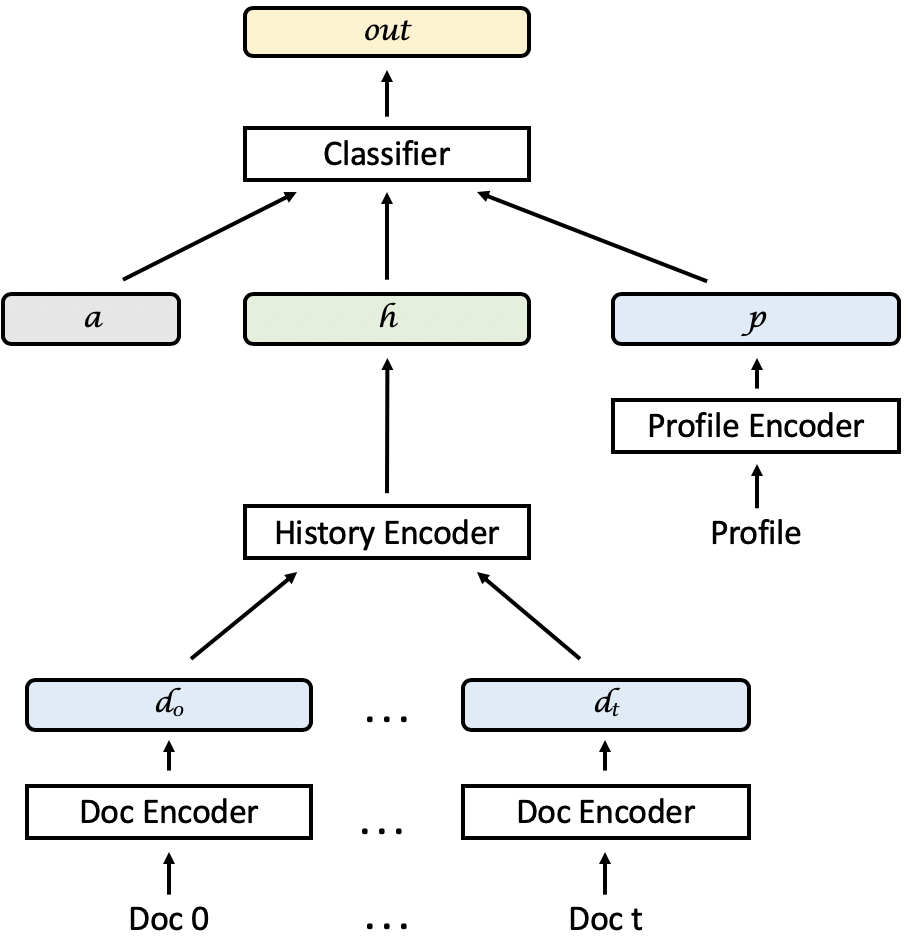}
\caption{Predictive model architecture.}
\label{fig:arch}
\end{figure}

We train a deep neural model, summarized in Figure \ref{fig:arch}, to take a user's \textbf{history}, \textbf{profile}, and \textbf{attributes}, and output a probability distribution over the set of $k_{act}$ clusters of human activities, indicating the likelihood that the user has reported to have performed an activity in each cluster. There are four major components of our network:
\begin{description}
\item [Document Encoder] This is applied to each of the $t$ documents in the history-- either an activity phrase or a full tweet. For document $i$ in $D$, it takes a sequence of token embeddings as input and produces a $dim_{d}$ dimensional vector, $\mathbf{d_i}$ as output.
\item [History Encoder] This layer takes the sequence $\{\mathbf{d_0},\ldots,\mathbf{d_t}\}$ as input and produces a single $dim_{H}$ dimensional vector, $\mathbf{h}$, as output, intended to represent high-level features extracted from the entire \textbf{history} of the user.
\item [Profile Encoder] Takes each token in the user's profile as input and produces a single $dim_{p}$ dimensional vector, $\mathbf{p}$ as output.

\item [Classifier] As input, this module takes the concatenation $\mathbf{a} \oplus \mathbf{h} \oplus \mathbf{p}$, where $\mathbf{a}$ is the predefined attribute vector associated with the user. Then, a prediction is made for each of the $k_{act}$ clusters, first applying softmax in order to obtain a probability distribution. We refer to the dimension of the output as $dim_o$.
\end{description}
For any of the three encoder layers, several layer types can be used, including recurrent, convolutional, or self-attention based \cite{transformer} layers. The classifier layer is the only layer that does not take a sequence as input and we implement it using a simple feed-forward multi-layer network containing $\ell_{c}$ layers with $h_{c}$ hidden units each. The network is trained with cross-entropy loss, which has been shown to perform competitively when optimizing for top-k classification tasks \cite{berrada2018smooth}.

\subsection{Incorporating Personal Values} \label{sec:values}

While the \textbf{attributes} vector $\mathbf{a}$ can be used to encode any information of interest about a user, we choose to experiment with the use of personal values because of their theoretical connection to human activities \cite{bardi2003values}. In order to get a representation of a user's values, we turn to the hierarchical personal values lexicon from \cite{wilson2018building}. In this lexicon, there are 50 value dimensions, represented as sets of words and phrases that characterize that value. Since users' profiles often contain value-related content, we use the Distributed Dictionary Representations (DDR) method \cite{garten2018dictionaries} to compute a score, $s_v$ for each value dimension, $v$, using cosine similarity as follows:
$$
s_v = \frac{R(profile) \cdot R(lexicon_v)}{||R(profile)|| || R(lexicon_v)||}\,,
$$
where $R(\cdot)$ is a representation of a set of vectors, which, for the DDR method, is defined as the mean vector of the set; $profile$ is a set of word embeddings, one for each token in the user's profile; and $lexicon_v$ is another set of word embeddings, one for each token in the lexicon for value dimension $v$. Finally, we set $\mathbf{a} = (s_0,\ldots,s_{dim_L})$ where $dim_L=50$, the number of value dimensions in the lexicon. 
Examples of profiles with high scores for sample value dimensions are shown in Table \ref{tab:values_in_profiles}.

 \begin{table}[htb]
    \centering
    \scalebox{0.9}{
    \begin{tabular}{c|p{5cm}}
    \hline 
        \textbf{Category} & \textbf{Top Scoring Profile} \\ \hline
        Family & a mother to my son \\ \hline
        Nature & Environment \& nat resource economist tweeting about climate change/risk, energy, environmental protection, green finance, commodities, data science, politics\\ \hline
        Work-Ethic & Football is like life - it requires perseverance, self-denial, hard work, sacrifice, dedication and respect for authority \\ \hline
        Religion & /Galatians 2:20/ I love our Lord Jesus Christ. \\ \hline
    \end{tabular}
    }
    \caption{Profiles scoring the highest for various values categories when  measured with the values lexicon. \vspace{-0.07in}}
    \label{tab:values_in_profiles}
\end{table}

\begin{table}[]
    \centering
    \scalebox{0.9}{
    \begin{tabular}{c|l}
    \hline 
        \textbf{Category} & \textbf{Activities in High Scoring Cluster} \\ \hline
                & give one's daughter a number of plants \\
        Family  & take one's family to the park \\
                & work in the garden with mom
                 \\ \hline
                & visit another castle \\
        Nature  & visit france \\
                & go on a fishing trip 
                 \\ \hline
                   & add another footnote to the dissertation \\
        Work-Ethic & file a complaint with the fcc \\
                   & write one's first novel by hand
                     \\ \hline
                 &  follow the rules \\
        Religion &            study really hard \\
                 & do a good deed\\ \hline
    \end{tabular}
    }
    \caption{Activity clusters associated with the highest scoring users for various values categories when measured with the values lexicon.}
    \label{tab:values_in_clusters}
\end{table}

Further, we explore the types of activity clusters that contain activities reported by users with high scores for various value dimensions. For a given value, we compute a score for each cluster $s^C_v$ by taking the average $s_v$ of all users who tweeted about doing activities in the cluster. For each value $v$, we can then rank all clusters by their $s^C_v$ score. Examples of those with the highest scores are presented in Table \ref{tab:values_in_clusters}. We observe that users whose profiles had high scores for Family were likely to report doing activities including  family members, those with high scores for Nature tweeted about travel, and those with high Work-Ethic scores reported performing writing related tasks.

\subsection{Evaluation}

We evaluate our activity prediction models using a number of metrics that consider not only the most likely cluster, but also the set of $k_{eval}$ most likely clusters. First, we evaluate the average per-class accuracy of the model's ability to rank $c_t$, the target cluster, within the top $k_{eval}$ clusters. These scores tell us how well the model is able to make predictions about the kinds of activities that each user is likely to do.

Second, we test how well the model is able to sort users by their likelihood of having reported to do an activity from a cluster. This average comparison rank (ACR) score is computed as follows: for each user in the test set, we sample $n$ other users who do not have the same activity label. Then, we use the probabilities assigned by the model to rank all $n+1$ users\footnote{We set $n=999$ in this study to achieve comparison samples of size $1000$.} by their likelihood of being assigned $c_t$, and the comparison rank score is the percentage of users who were ranked ahead of the target user (lower is better). We then average this comparison rank across all users in the test set to get the ACR. The ACR score tells us how well the model is able to find a rank users based on their likelihood of writing about doing a given activity, which could be useful for finding, e.g., the users who are most likely to claim that they ``purchased some pants'' or least likely to mention that they ``went to the gym'' in the future.


\section{Experiments and Results}


We split our data at the user-level, and from our set of valid users we use 200,000 instances for training data, 10,000 as test data, and the rest as our validation set.

\begin{table*}[t]
\centering
\scalebox{0.9}{
\begin{tabular}{l|cccccc||c}
\hline 
\textbf{$k_{eval}$} & 1 & 2 & 3 & 5 & 10 & 25 & \textbf{ACR}\\ \hline
$\mathbf{full_T}$ & 2.54 & 5.04 & 7.01 & 13.14 & 24.49 & 55.36 & 46.22 \\
$-a$ & 2.11 & 5.05 & 7.91 & 13.58 & 23.29 & 54.85 & 46.12 \\
$-p$ & 3.20 & 6.47 & 9.08 & 14.70 & 27.52 & 60.26 & 42.24 \\
$-a,p$ &\textbf{ 4.29} & \textbf{7.76} & \textbf{10.67} &\textbf{ 15.92} & \textbf{29.12} & \textbf{61.03} & \textbf{41.51} \\ \hline
$\mathbf{full_A}$ & 2.13 & 4.46 & 7.12 & 11.44 & 22.49 & 55.05 & 47.40 \\
$-a$ & 2.60 & 4.55 & 7.35 & 12.26 & 23.37 & 54.73 & 46.17 \\
$-p$ & 2.75 & 4.84 & 7.56 & 12.00 & 25.25 & 55.36 & 46.23 \\
$-a,p$ & 3.75 & 6.79 & 9.73 & 15.47 & 28.22 & 60.87 & 42.70 \\ \hline
$-h$ & 2.02 & 4.13 & 6.67 & 11.61 & 23.43 & 53.38 & 47.98 \\
$-a,h$ & 1.68 & 4.55 & 7.61 & 11.49 & 23.41 & 52.97 & 47.83 \\
$-p,h$ & 2.29 & 3.61 & 4.88 & 9.22 & 20.48 & 51.25 & 49.28 \\ \hline
\textbf{rand} & 2.00 & 4.00 & 6.00 & 10.00 & 20.00 & 50.00 & 50.00 \\ \hline
\end{tabular}
}
\caption{Per-class accuracy (\%) @ $k_{eval}$ and ACR scores for the 50-class prediction task. Note that removing $h$ from either $\mathbf{full_T}$ or $\mathbf{full_A}$ gives the same model. For ACR only, lower is better. \vspace{-0.1in}}
\label{tab:50acc_at_k}
\end{table*}

For the document encoder and profile encoder we use Bi-LSTMs with max pooling \cite{conneau2017supervised}, with $dim_d=128$ and $dim_p=128$. For the history encoder, we empirically found that single mean pooling layer over the set of all document embeddings outperformed other more complicated architectures, and so that is what we use in our experiments. Finally, the classifier is a 3-layer feed-forward network with and $dim_c=512$ for the hidden layers, followed by a softmax over the $dim_o$-dimensional output. We use Adam \cite{kingma2014adam} as our optimizer, set the maximum number of epochs to 100, and shuffle the order of the training data at each epoch. During each training step, we represent each user's history as a new random sample of $max\_sample\_docs=100$ documents\footnote{We empirically found that increasing this value beyond 100 had little effect on the development accuracy.} if there are more than $max\_sample\_docs$ documents available for the user, and we use a batch size of 32 users. Since there is a class imbalance in our data, we use sample weighting in order to prevent the model from converging to a solution that simply predicts the most common classes present in the training data. Each sample is weighted according to its class, $c$, using the following formula:
$$
w_c = \frac{N}{count(c) * dim_o}
$$
where $count(c)$ is the number of training instances belonging to class $c$. We evaluate our model on the development data after each epoch and save the model with the highest per-class accuracy. Finally, we compute the results on the test data using this model, and report these results.


We test several configurations of our model. We use the complete model described in section \ref{sec:arch} using either the set of \textit{additional tweets} written by a user as their \textbf{history} ($\mathbf{full_T}$), or only the set of \textit{additional activities} contained in those tweets ($\mathbf{full_A}$). Then, to test the effect of the various model components, we systematically ablate the attributes vector input $a$, the profile text (and subsequently, the Profile Encoder layer) $p$, and the set of documents, D, comprising the history along with the Document and History Encoders, thereby removing the $h$ vector as input to the classifier. We also explore removing pairs of these inputs at the same time. To contextualize the results, we also include the theoretical scores achieved by random guessing, labeled as \textbf{rand}.\footnote{For the evaluation metrics considered in this paper, random guessing is as strong or stronger than a ``most frequent class'' baseline, so we do not report it.}


We consider two variations on our dataset: the first is a simplified, 50-class classification problem. We choose the 50 most common clusters out of our full set of $k_{act}=1024$ and only make predictions about users who have reportedly performed an activity in one of these clusters. The second variation uses the entire dataset, but rather than making predictions about all $k_{act}$ classes, we only make fine-grained predictions about those classes for which $count(c) \geq minCount$. We do this under the assumption that training an adequate classifier for a given class requires at least $minCount$ examples. All classes for which $count(c) < minCount$ are assigned an ``other'' label. In this way, we still make a prediction for every instance in the dataset, but we avoid allowing the model to try to fit to a huge landscape of outputs when the training data for some of these outputs is insufficient. By setting $minCount$ to 100, we are left with 805 out of 1024 classes, and an 806th ``other'' class for our 806-class setup. Note that this version  includes all activities from all 1024 clusters, it is just that the smallest clusters are grouped together with the ``other'' label.

\begin{table*}[ht]
\centering
\scalebox{0.9}{
\begin{tabular}{l|ccccccccccc||c}
\hline 
\textbf{$k_{eval}$} & 1 & 2 & 3 & 5 & 10 & 25 & 50 & 75 & 100 & 200 & 300 & \textbf{ACR} \\ \hline
$\mathbf{full_T}$ & 0.15 & 0.36 & 0.61 & 0.97 & 1.91 & 4.65 & 8.66 & 12.24 & 16.15 & 30.69 & 43.96 & 44.10 \\
$-a$& 0.32 & 0.61 & 0.98 & 1.39 & 2.96 & 5.99 & 10.21 & 14.61 & 18.95 & 35.19 & 49.26 & 42.61 \\
$-p$ & \textbf{0.45} & \textbf{1.02 }& \textbf{1.37} & \textbf{1.96} & \textbf{3.38} & \textbf{7.41} & 12.71 & 17.17 & 21.60 & 37.53 & 51.11 & 41.14 \\
$-a,p$ & 0.41 & 0.70 & 1.10 & 1.66 & 3.03 & 6.88 & \textbf{12.89} & \textbf{17.86} & \textbf{22.76} & \textbf{38.61} & \textbf{52.38} & \textbf{40.82} \\ \hline
$\mathbf{full_A}$ & 0.29 & 0.41 & 0.72 & 1.04 & 2.05 & 4.50 & 8.50 & 12.14 & 15.48 & 30.04 & 44.24 & 45.98 \\
$-a$ & 0.24 & 0.44 & 0.75 & 1.02 & 2.02 & 4.62 & 8.70 & 12.19 & 15.56 & 30.18 & 43.34 & 45.99 \\
$-p$ & 0.23 & 0.46 & 0.66 & 1.13 & 2.29 & 5.27 & 9.66 & 14.33 & 18.75 & 34.00 & 47.71 & 42.64 \\
$-a,p$ & 0.26 & 0.47 & 0.83 & 1.35 & 2.24 & 4.61 & 8.90 & 13.24 & 16.80 & 31.29 & 45.11 & 44.56 \\ 
\hline
$-h$ & 0.10 & 0.28 & 0.44 & 0.73 & 1.37 & 4.08 & 7.60 & 10.96 & 14.28 & 27.60 & 40.77 & 47.94 \\
$-a,h$ & 0.10 & 0.36 & 0.53 & 1.00 & 1.85 & 4.64 & 8.58 & 12.57 & 16.23 & 29.31 & 41.57 & 46.94 \\
$-p,h$ & 0.10 & 0.23 & 0.41 & 0.68 & 1.49 & 3.72 & 7.12 & 10.46 & 13.65 & 26.90 & 39.93 & 48.15 \\ \hline
\textbf{rand} & 0.12 & 0.25 & 0.37 & 0.62 & 1.24 & 2.98 & 6.34 & 9.19 & 12.54 & 26.21 & 36.77 & 50.00  \\ \hline
\end{tabular}
}
\caption{Per-class accuracy (\%) @ $k_{eval}$ and ACR scores for the 806-class prediction task. Note that removing $h$ from either $\mathbf{full_T}$ or $\mathbf{full_A}$ gives the same model. For ACR only, lower is better. \vspace{-0.1in}}
\label{tab:806acc_at_k}
\end{table*}

While our models are able to make predictions indicating that learning has taken place, it is clear that this prediction task is difficult. In the 50-class setup, the $\mathbf{full_T} -a,p$ model consistently had the strongest average per-class accuracy for all values of $k_{eval}$ and the lowest (best) ACR score (Table \ref{tab:50acc_at_k}). The $\mathbf{full_A} -a,p$ model performed nearly as well, showing that using only the human-activity relevant content from a user's \textbf{history} gives similar results to using the full set of content available. When including the \textbf{attributes} and \textbf{profile} for a user, the model typically overfits quickly and generalization deteriorates.

In the 806-class version of the task, we observe the effects of including a larger range of activities, including many that do not appear as often as others in the training data (Table \ref{tab:806acc_at_k}). This version of the task also simulates a more realistic scenario, since predictions can be made for the ``other'' class when the model does to expect the user to claim to do an activity from \textit{any} of the known clusters. In this setting, we see that the $\mathbf{full_T}-p$ model works well for $k_{eval}\leq 25$, suggesting that the use of the $attribute$ vectors helps, especially when predicting the correct cluster within the top 25 is important. For $k_{eval} \geq 50$, the same $\mathbf{full_T}-a,p$ model that worked best in the 50-class setup again outperforms the others. Here, in contrast to the 50-class setting, using the full set of tweets usually performs better than focusing only on the human activity content. Interestingly, the best ACR scores are even lower in the 806-class setup, showing that it is just as easy to rank users by their likelihood of writing about an activity, even when considering many more activity clusters.

\section{Conclusions} 

In this paper, we  addressed the task of predicting human activities from user-generated content. We collected a large Twitter dataset consisting of posts from more than 200,000 users mentioning at least one of the nearly 30,000 everyday activities that we explored. 
Using sentence embedding models, we projected activity instances into a vector space and perform clustering in order to learn about the high-level groups of behaviors that are commonly mentioned online. We trained predictive models to make inferences about the likelihood that a user had reported to have done activities across the range of clusters that we discovered, and found that these models were able to achieve results significantly higher than random guessing baselines for the metrics that we consider. While the overall prediction scores are not very high, the models that we  trained do show that they are able to generalize findings from one set of users to another. This is evidence that the task is feasible, but very difficult, and it could benefit from further investigation.

We make the activity clusters, models, and code for the prediction task available at http://lit.eecs.umich.edu/downloads.html



\section*{Acknowledgments}


This research was supported in part through computational resources and services provided by the Advanced Research Computing at the University of Michigan. This material is based in part upon work supported by the Michigan Institute for Data Science, by the National Science Foundation (grant \#1815291), by the John Templeton Foundation (grant \#61156), and by DARPA (grant \#HR001117S0026-AIDA-FP-045). Any opinions, findings, and conclusions or recommendations expressed in this material are those of the author and do not necessarily reflect the views of the Michigan Institute for Data Science, the National Science Foundation, the John Templeton Foundation, or DARPA. Many thanks to the anonymous reviewers who provided helpful feedback.

\bibliographystyle{acl_natbib}
\bibliography{1_thebibliography}

\end{document}